\documentclass{ifacconf}

\usepackage{graphicx}      
\usepackage{amssymb}
\usepackage{mathalpha}
\usepackage{amsmath}
\usepackage{multirow}
\usepackage{xcolor}
\usepackage{soul}

\usepackage{natbib}        

\begin{document}
\begin{frontmatter}

\title{ViTA-Seg: Vision Transformer for Amodal Segmentation in Robotics} 


\author[First]{Donato Caramia} 
\author[Second]{Florian T. Pokorny}
\author[Third]{Giuseppe Triggiani}
\author[Third]{Denis Ruffino}
\author[First]{David Naso}
\author[First]{Paolo Roberto Massenio} 

\address[First]{Department of Electrical and Information Engineering (DEI), 
 \\ Polytechnic University of Bari, 70126, Bari, Italy, \\ (e-mail: d.caramia2@phd.poliba.it, \{paoloroberto.massenio, david.naso\} @poliba.it).}
\address[Second]{Division of Robotics, Perception, and Learning (RPL),\\ KTH Royal Institute of Technology, 114 28, Stockholm, Sweden, (e-mail: fpokorny@kth.se)}
\address[Third]{AROL S.p.A., 14053, Canelli, Italy, (e-mail: \{giuseppe.triggiani, denis.ruffino\} @arol.com)}

\begin{abstract}  
Occlusions in robotic bin picking compromise accurate and reliable grasp planning. We present ViTA-Seg, a class-agnostic Vision Transformer framework for real-time amodal segmentation that leverages global attention to recover complete object masks, including hidden regions. We proposte two architectures: a) Single-Head for amodal mask prediction; b) Dual-Head for amodal and occluded mask prediction. We also introduce ViTA-SimData, a photo-realistic synthetic dataset tailored to industrial bin-picking scenario. Extensive experiments on two amodal benchmarks, COOCA and KINS, demonstrate that ViTA-Seg Dual Head achieves strong amodal and occlusion segmentation accuracy with computational efficiency, enabling robust, real-time robotic manipulation.
\end{abstract}


\end{frontmatter}

\section{Introduction}
Robotic bin picking is a fundamental task in industrial automation, where a vision-guided robot must detect, localize, and grasp objects from a container filled with randomly stacked items. Despite significant progress in perception and manipulation, cluttered bins remain challenging due to occlusions and complex object interactions (\cite{zhang2022regrad}). 
In this context, amodal segmentation is the task of predicting the complete mask of an object, including both its visible and occluded parts. By recognizing both occluding and occluded objects, the robot can prioritize which object to pick first and avoid failed grasps or collisions, reducing cycle time and increasing throughput (\cite{gilles2023metagraspnetv2}). 



Amodal segmentation is predominantly addressed using deep-learning models that infer the hidden portions of objects from visual cues (\cite{Gao2023C2FSeg}). However, learning such models remains difficult because most large-scale datasets (e.g., COCO, KITTI) provide only modal masks, leaving occluded regions unlabeled. Collecting amodal annotations, i.e., pixel-wise labels for visible, occluded, and complete object masks, is labor-intensive and hard to scale. As a result, existing methods generalize poorly to unseen shapes and struggle in class-agnostic settings, where objects must be segmented regardless of category.

\begin{figure}
    \centering
    \includegraphics[scale = 0.8]{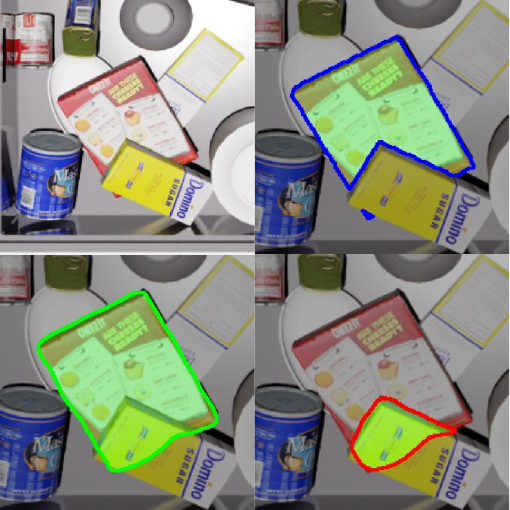}
    \caption{Sample of ViTA-SimData (top-left), visible mask (top-right), predicted amodal (bottom-left) and occluded mask (bottom-right) of ViTA-Seg Dual Head.}
    \label{samples}
\end{figure}

To address data and generalization challenges, early work extended Mask R-CNN to predict visible, occluded, and full amodal masks, as in \cite{Follmann2018SeeInvisible}. These single-stage convolutional neural network (CNN) models learn pixel-wise amodal labels, but they rely on expensive dense annotations and struggle to generalize to novel object shapes. 
UOAIS-Net (\cite{Back2021UOAIS}), also built upon Mask R-CNN, predicts object class, bounding-box, visible, amodal, and occlusion masks from RGB-D input. Although it improves occlusion reasoning, it depends on depth sensing and a multi-stage pipeline that slows inference. Like other CNN-based methods, it still struggles to generalize to unseen geometries, limiting its suitability for cluttered bin-picking scenarios.

Recent advances in Vision Transformers (ViT) show strong performance thanks to their ability to capture long-range dependencies and global context, enabling better reasoning about object relationships and spatial structure (\cite{dosovitskiy2020image}). Their self-attention mechanisms and large-scale pretraining further improve generalization to unseen categories (\cite{mae}), making ViTs effective backbones for amodal segmentation in industrial robotic perception.  AISFormer (\cite{tran2022aisformer}) employs a transformer decoder to predict occluder, visible, amodal, and occluded masks from RGB images. However, decoding multiple object instances and generating four masks introduces significant computational cost. Other approaches incorporate global shape priors to guide the completion of occluded regions. ShapeFormer (\cite{Tran2024ShapeFormer}) pairs a visible-to-amodal transformer with a generative autoencoder 
whose latent space compresses object shapes into compact codes. These global shape cues improve amodal completion, but they require a sufficiently rich prior database, which is difficult to build in class-agnostic settings. Diffusion models improve geometric coherence by iteratively denoising latent space to reconstruct complete shapes. \cite{Zhan2024AmodalWild} uses Stable Diffusion for full-shape prediction, while AISDiff (\cite{Tran2024AISDiff}) refines amodal masks with a diffusion-based shape prior. These methods produce high-quality completions but require heavy pretrained models and slow inference, limiting industrial use. In contrast, C2F-Seg (\cite{Gao2023C2FSeg}) offers a more lightweight, class-agnostic alternative. It encodes the region of interest and visible mask into a latent space, predicts a coarse amodal token sequence via a mask-and-predict module (\cite{chang2022maskgit}), and refines it with a compact decoder. While effective, its iterative prediction process still introduces non-negligible latency.

Despite recent progress, most existing amodal segmentation methods are designed and benchmarked on general-purpose datasets (e.g., COCOA \cite{cocoa}, KINS \cite{kins}), with limited consideration for clutter, high occlusion, and class-agnostic requirements typical of industrial bin-picking. Current approaches suffer from three main limitations: (i) the scarcity of labeled amodal data in industrial settings, due to the time-consuming annotation process; (ii) poor generalization to novel object shapes, especially in class-agnostic environments; and (iii) high computational cost or inference latency that restricts real-time applicability. These gaps motivate the need for a fast, class-agnostic framework capable of global reasoning over cluttered scenes to predict both visible and occluded regions. The main contributions of this paper are as follows:
\begin{itemize}
   \item A novel ViT-based amodal segmentation model (referred to as ViTA-Seg) designed to predict complete object masks, including both visible and hidden regions, in cluttered bin-picking scenes. ViTA-Seg leverages the global contextual reasoning of ViT to accurately segment both occluding and occluded objects. We introduce two architectures: ViTA-Seg Single Head, which predicts only the amodal mask, and ViTA-Seg Dual Head, which predicts both amodal and occluded masks. Without relying on latent-space representations, our models achieve superior accuracy and reduced inference time compared to state-of-the-art class-agnostic methods such as C2F-Seg.
    \item A new synthetic amodal dataset, ViTA-SimData (see Fig. \ref{samples}), specifically tailored to industrial bin-picking scenarios. The dataset includes 583 photo-realistic images and 9,180 object instances with detailed annotations (bounding box, amodal, visible, and occlusion masks), enabling training and evaluation under realistic cluttered and occluded conditions.
    \item Extensive experiments on real-world benchmarks such as COCOA (\cite{cocoa}) and KINS (\cite{kins}) confirm that ViTA-Seg outperforms prior state-of-the-art models in both accuracy and computational efficiency. These results demonstrate not only its suitability for real-time bin-picking perception but also its applicability to domains beyond bin picking.
\end{itemize}

The remainder of this paper is organized as follows. Section 2 describes the proposed architectures. Section 3 presents the results, including the datasets, training setup, and baseline comparisons. Concluding remarks are reported in Section 4. 

\begin{figure*}
    \centering
    \includegraphics[scale = 0.7]{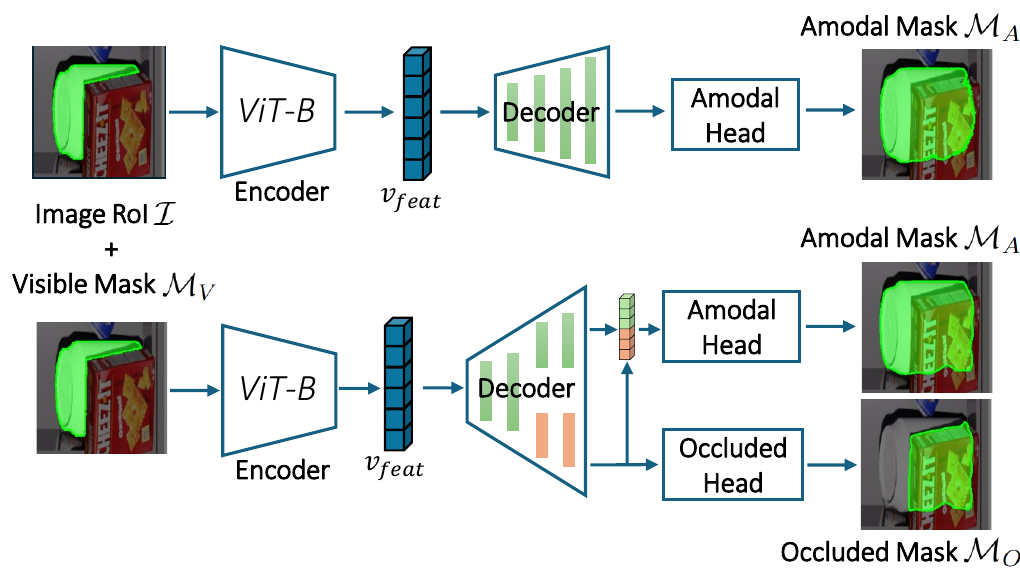}
    \caption{The proposed ViTA-Seg architectures. ViTA-Seg Single Head (top) and ViTA-Seg Dual Head (bottom).}
    \label{ViTAs}
\end{figure*}

\section{The Proposed ViTA-Seg architecture}

In this paper, we propose two architectures: ViTA-Seg Single Head, which predicts only the amodal mask, and ViTA-Seg Dual Head, which predicts both amodal and occluded masks (Fig. \ref{ViTAs}). Both models take two complementary inputs: (1) a rescaled $224\times224$ pixel Region of Interest (RoI), $\mathcal{I}$, extracted by cropping the original image around the visible-mask bounding box, $\mathcal{M}_V$, enlarged by 20\% to preserve local context around the object; and (2) the visible mask $\mathcal{M}_V$ itself. 
In practice, $\mathcal{M}_V$ is obtained from an upstream detector such as GroundingDINO (\cite{liu2024grounding}), which outputs a bounding box and visible segmentation, $\mathcal{M}_V$, without relying on predefined classes. Since no semantic labels are used, the task remains class-agnostic: GroundingDINO localizes the object, and ViTA-Seg predicts its amodal and occluded components. 

ViTA-Seg follows an encoder–decoder design. The encoder extracts visual features, $v_{feat}$, that capture both low-level appearance and high-level structural information. The decoder then upsamples and refines these features to produce the final $224\times224$ pixel masks: the amodal mask $\mathcal{M}_A$ for the Single-Head model, and both $\mathcal{M}_A$ and the occluded mask $\mathcal{M}_O$ for the Dual-Head variant. The encoder and decoder modules are described in the following subsections.

\subsection{Encoder}

The encoder is built upon a ViT backbone with the base configuration (ViT-B) pre-trained with Masked Autoencoders (MAE), which are particularly effective at reconstructing hidden image regions (\cite{mae}). MAE pre-training randomly masks a large portion of the image and trains the model to recover the missing pixels, thus enabling feature learning for occlusion-aware tasks. The encoder takes a 4-channel input constructed by concatenating the 3-channel RGB $\mathcal{I}$ with the single-channel of $\mathcal{M}_V$. This concatenation allows to simultaneously process visual appearance information and explicit knowledge about which regions are currently visible. To adapt the pre-trained ViT weights from 3 to 4 input channels, we initialize the additional channel's projection weights as the average of the RGB channel weights.
Then, the ViT-B encoder outputs a 768-dimensional feature vector:
\begin{equation}
    v_{feat} = ViT_B( \mathcal{I},\mathcal{M}_V).
\end{equation}

\subsection{Decoder}
The decoder uses convolutional layers to upsample the compressed features and produce segmentation masks at the $224\times224$ pixel resolution.


\subsubsection{Single Head.} 

The single-head architecture maps $v_{feat}$ directly to the amodal mask $\mathcal{M}_A$. The 768-dimensional encoder output is passed through a sequence of 4 decoder blocks that progressively increase spatial resolution while reducing channel depth, following the transformation sequence $[768,1,1] \rightarrow [512,28,28] \rightarrow [256,56,56] \rightarrow [128,112,112] \rightarrow [64, 224,224]$. Each block uses transposed convolutions for upsampling, followed by standard convolutions, batch normalization, and ReLU activations. Finally, the decoder is followed by a two-layer convolutional head that progressively reduces the channel depth from $[64,224,224] \rightarrow [32,224,224] \rightarrow [1,224,224]$, after which a sigmoid activation generates the final binary mask.

The loss function for the single-head architecture is the Binary Cross-Entropy (BCE) loss (\cite{Gao2023C2FSeg}), computed as:
\begin{equation}
\mathcal{L}_{single} = BCE(\mathcal{M}_A, \mathcal{M}_A^{gt}),
\end{equation}
where $\mathcal{M}_A^{gt}$ is the ground-truth of amodal mask.

\subsubsection{Dual Head.}
Following the multi-head strategy explored in \cite{Gao2023C2FSeg} and \cite{Back2021UOAIS}, we design a dual-head architecture that incorporates cross-task collaboration between amodal and occluded prediction. The decoder begins with 2 shared layers that extract features useful to both tasks, with transformation sequence $[768,1,1] \rightarrow [512,28,28] \rightarrow [256,56,56]$. Then, the architecture splits into two task-specific branches, each starting from a $[128,56,56]$ feature map. Both branches contain 2 dedicated decoder layers that upsample and refine the representation with the dimensional progression $[128,56,56] \rightarrow [64,112,112] \rightarrow [32,224,224]$. Cross-task collaboration is introduced at the final stage: the $32$-channel features from the occluded branch are concatenated with the $32$-channel features from the amodal branch, allowing the amodal path to benefit from explicit cues about hidden regions while keeping the occluded path independent.

The architecture is completed by two prediction heads. The amodal head consists of two convolutional layers that progressively reduce the channels from $[64,224,224]$ to $[32,224,224]$ and finally to $[1,224,224]$, after which a sigmoid activation produces the amodal mask $\mathcal{M}_A$. In parallel, the occluded head applies a single convolutional layer that maps the $[32,224,224]$ channels to a $[1,224,224]$ output, followed by a sigmoid activation to generate the occluded mask $\mathcal{M}_O$.

The dual-head architecture is trained by supervising both outputs jointly. The amodal and occluded heads use binary cross-entropy (BCE) losses:
\begin{equation}
\mathcal{L}_{A} = BCE(\mathcal{M}_A, \mathcal{M}_A^{gt}), \qquad
\mathcal{L}_{O} = BCE(\mathcal{M}_O, \mathcal{M}_O^{gt}),
\end{equation}
where $\mathcal{M}_O^{gt}$ is the ground-truth occluded mask. The total loss is a weighted sum:
\begin{equation}
\mathcal{L}_{total} = \lambda_{A}\mathcal{L}_{A} + \lambda_{O}\mathcal{L}_{O},
\label{eq_total_loss}
\end{equation}
with $\lambda_{A}$ and $\lambda_{O}$ as design parameters.

\section{Results}

In this section, we evaluate the performance of ViTA-Seg models across multiple datasets and under different occlusion conditions. We first describe the datasets and training setup, then present quantitative and qualitative results comparing ViTA-Seg with a state-of-the-art baseline.

\subsection{Employed Datasets}

We evaluate our approach on two established amodal benchmarks and on a new synthetic dataset tailored to industrial bin-picking scenarios, as detailed below.

\subsubsection{1) COCOA (\cite{cocoa}).}
Derived from MS COCO, COCOA is a standard benchmark for amodal segmentation. It contains real-world images with diverse object categories and complex occlusions, providing manually annotated visible and amodal masks. The dataset includes 2,476 training images and 1,223 validation images. Since each image may contain multiple objects, the number of annotation triplets $(\mathcal{M}_V,\mathcal{M}_A,\mathcal{M}_O)$ is higher than the number of images: the training set contains 6,763 annotated instances, while the validation set contains 3,799.

\subsubsection{2) KINS (\cite{kins}).}
KINS extends KITTI with amodal annotations for autonomous driving, focusing on traffic participants and road-side objects, featuring challenging occlusions common in driving environments. It contains 7,474 training images and 7,517 validation images, yielding 95,456 annotated instances in the training set and 92,618 in the validation set.

\subsubsection{3) ViTA-SimData (synthetic).}
ViTA-SimData is a synthetic dataset that we developed specifically for the bin-picking scenario. Each scene is generated in Isaac Sim (\cite{isaac}) by randomly selecting between 15 and 20 objects from ShapeNet (\cite{shapenet}) and NVIDIA Omniverse Assets (\cite{nvidiaasset}), including everyday items such as cans and bottles, and dropping them into a bin using physics simulation to create realistic clutter and occlusions. A virtual top-view camera captures the RGB image of each scene. For each object instance, the pipeline automatically produces two masks: the visible mask $\mathcal{M}_V$ and the amodal mask $\mathcal{M}_A$, while the occluded mask is obtained as $\mathcal{M}_O = \mathcal{M}_A - \mathcal{M}_V$.
The dataset contains 583 images in total, 383 for training and 200 for validation, corresponding to 5,364 annotated instances in the training split and 3,816 in the validation split. To improve generalization, training and validation use disjoint object sets: no 3D model appearing in the training set is used in validation. Examples from this dataset are in Fig. \ref{samples} and Fig. \ref{vita-vs-world}. 

Finally, we group samples into three levels of occlusion-rate ($occ_R$): low (0–20\%), medium (20–50\%), and high (50–100\%).  Table~\ref{tabdata} reports the percentage of samples falling into each category for all datasets.

\subsection{Model Training}

ViTA-Seg is implemented in PyTorch and trained using the AdamW optimizer with a learning rate of $5\times10^{-6}$, $\beta_1=0.9$, and $\beta_2=0.999$. Both the single-head and dual-head models are trained with a fixed batch size of 8 for each dataset. Since ViTA-SimData is synthetically generated, we adopt a transfer-learning strategy: the models are first trained from scratch on COCOA, and the resulting weights are then used as initialization for training on ViTA-SimData. This provides meaningful priors on object appearance and spatial structure learned from real images, helping mitigate the domain gap between synthetic and real data. We do not use KINS for this initialization because its autonomous-driving domain differs substantially from bin-picking and would introduce domain-specific biases; additionally, its occlusion-rate distribution (especially at low occlusion levels) differs markedly from that of ViTA-SimData, while COCOA is more closely aligned (see Tab.~\ref{tabdata}). Nevertheless, KINS is still included in the evaluation to assess the prediction capabilities of the proposed models. For the total loss, we set $\lambda_{A} = 1.0$ and $\lambda_{O} = 0.25$. Training and inference are performed on a workstation equipped with an NVIDIA GeForce RTX 4090, i9-14900KF CPU, and 128~GB RAM.

\subsection{Results}

We compare ViTA-Seg against C2F-Seg, which, to the best of our knowledge, is the only existing amodal segmentation method that operates in a class-agnostic setting while delivering strong performance. Model performance is evaluated using standard amodal segmentation metrics. We compute Intersection over Union (IoU) for the amodal ($mIoU_A$), visible ($mIoU_V$), and occluded ($mIoU_O$) masks (\cite{yao2022self, kins, tangemannunsupervised}). We additionally measure the inference time ($t_{inf}$), averaged over all validation samples. For a fair comparison across datasets, all three models (ViTA-Seg with single and double head, and C2F-Seg) are trained from scratch on KINS and COCOA when evaluated on those datasets. For ViTA-SimData, we apply the transfer-learning strategy described before: both ViTA-Seg and C2F-Seg are first trained on COCOA, and the resulting weights are used to initialize training on ViTA-SimData.

\begin{table}[t]
\centering
\resizebox{\columnwidth}{!}{%
\begin{tabular}{cccc}\hline  
\multirow{2}{*}{Dataset} & \multicolumn{3}{c}{object occlusion distribution} \\
& \multicolumn{1}{c}{\textit{$occ_R$-low} } & \multicolumn{1}{c}{\textit{$occ_R$-medium}} & \multicolumn{1}{c}{\textit{$occ_R$-high}} \\ \hline \hline
KINS & 23.08 & \textbf{45.12} & \textbf{31.80} \\
COCOA & \textbf{58.88} & 29.96 & 11.16 \\
\begin{tabular}[c]{@{}c@{}}ViTA-SimData\end{tabular} & 43.50 & 33.69 & 22.81 \\\hline                 
\end{tabular}%
}
\caption{Distribution of object instances for occlusion-rates on the three datasets.}
\label{tabdata}
\end{table}

Table~\ref{tab1} summarizes the validation results across all datasets. The proposed Dual-Head variant consistently achieves state-of-the-art performance. As expected, avoiding both the iterative mask-and-predict process and the latent-space encoding of $\mathcal{I}$ and $\mathcal{M}_V$ used in C2F-Seg makes our models substantially faster, approximately 12 times in practice. ViTA-Seg Dual Head outperforms both the Single-Head version and C2F-Seg in predicting the amodal and occluded masks. Compared to C2F-Seg, the addition of the occlusion head improves $mIoU_A$ and $mIoU_O$ by 6 and 13 points on COCOA, by 4 and 6 points on KINS, and by 6 and 5 points on ViTA-SimData, respectively.

\begin{table*}[]
\centering
\resizebox{\textwidth}{!}{%
\begin{tabular}{c||ccc||ccc||ccc||c} \hline
\multirow{2}{*}{Model} & \multicolumn{3}{c||}{KINS} & \multicolumn{3}{c||}{COCOA} & \multicolumn{3}{c||}{ViTA-SimData} & \multirow{2}{*}{$t_{inf}$}     \\
& $mIoU_A$ & $mIoU_V$ & $mIoU_O$ & $mIoU_A$ & $mIoU_V$ & $mIoU_O$ & $mIoU_A$ & $mIoU_V$ & $mIoU_O$ \\ \hline
C2F-Seg & 87.89 & 72.12 & 57.30 & 87.15 & 92.52 & 36.69 & 85.17 & 90.67 & 53.28 & 113.77  \\
ViTA-Seg Single-Head & 89.94 & \textbf{98.84} & 60.05 & 91.57 & \textbf{99.63} & 27.65 & 88.31 & 91.01 & 49.33 & \textbf{8.54}  \\
ViTA-Seg Dual-Head & \textbf{91.12} & 98.65 & \textbf{63.48} & \textbf{93.70} & 99.38 & \textbf{49.88} & \textbf{91.09} & \textbf{91.48} & \textbf{58.65} & 8.96 \\ 
\hline 
\end{tabular} 
}
\caption{Performance comparison of ViTA-Seg and C2F-Seg models on the validation datasets.}
\label{tab1}
\end{table*}

\begin{figure*}
    \centering
    \includegraphics[scale = 1]{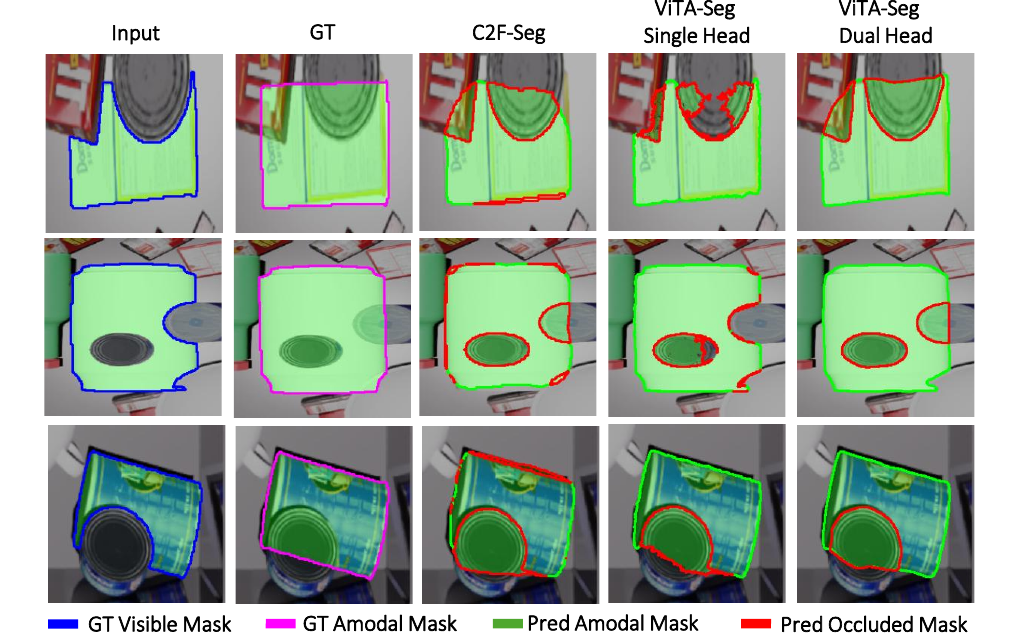}
    \caption{The qualitative results estimated by C2F-Seg, ViTA-Seg Single Head and Dual Head on ViTA-SimData.}
    \label{vita-vs-world}
\end{figure*}
\begin{figure*}
    \centering
    \includegraphics[scale = 1]{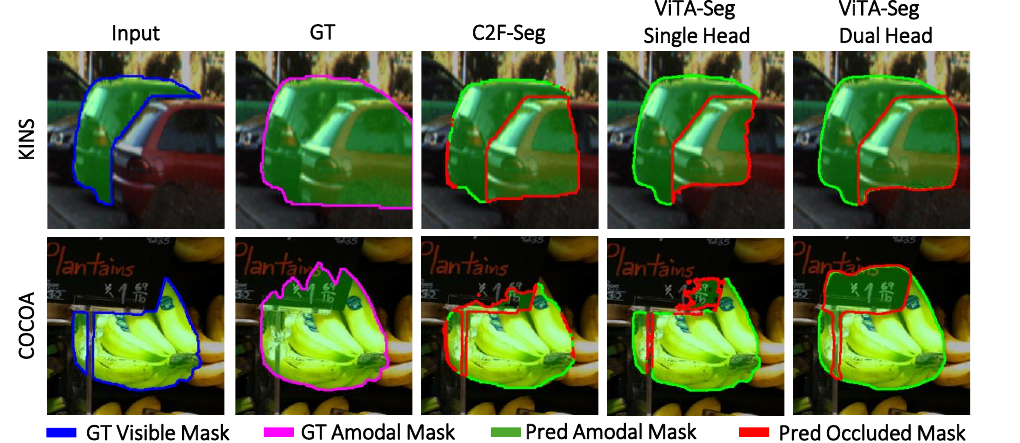}
    \caption{The qualitative results estimated by C2F-Seg, ViTA-Seg Single Head and Dual Head on KINS and COCOA.}
    \label{vita-vs-world-pt2}
\end{figure*}

As shown in Tab. \ref{occtab}, all models demonstrate improved accuracy in predicting medium and high occluded objects. This performance narrows the $mIoU$ gap between various models, given that ViTA-SimData possesses a distinct $occ_R$ distribution compared to both COCOA and KINS (Tab. \ref{tabdata}). However, the $mIoU_V$ achieved on ViTA-SimData has decreased if compared with the results on KINS and COCOA; as the dataset complexity increases, models are driven to prioritize the prediction of complex occluded regions at the expense of visible parts. This trade-off is visible in the qualitative results presented in Fig. \ref{vita-vs-world} and Fig. \ref{vita-vs-world-pt2}. While ViTA-Seg Dual Head predicts more occluded regions than other methods, it also (along with ViTA-Seg Single Head) adheres more closely to the visible mask. In contrast, C2F-Seg tends to hallucinate around the occluded mask (indicated by the red lines in Column 3 of Fig. \ref{vita-vs-world} and Fig. \ref{vita-vs-world-pt2}), partially forgetting the input mask to prioritize the amodal result.

\begin{figure}
    \centering
    \includegraphics[scale = 1]{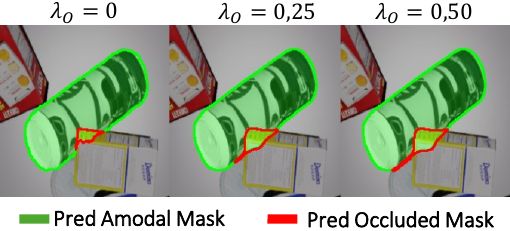}
    \caption{Predictions for different setting of $\lambda_O$.}
    \label{imgL}
\end{figure}

Finally, to evaluate the impact of the Occluded Head loss function, we analyze the performance of ViTA-Seg Dual Head across different weights $\lambda_O$ (cfr. \eqref{eq_total_loss}) in Tab. \ref{lambda_O}. The choice of $\lambda_O$ proves critical for balancing the total loss function. While increasing $\lambda_O$ to $0.50$ maximizes the occluded performance ($mIoU_O = 58.98$), it leads to a slight degradation in the overall amodal accuracy ($mIoU_A$). This trade-off occurs because assigning greater weight to the occluded head shifts the model's focus toward the occluded regions $\mathcal{M}_O$, which, by definition, exclude the visible mask ($\mathcal{M}_V \cap \mathcal{M}_O = \emptyset$), while $\mathcal{M}_V \subset \mathcal{M}_A$. Therefore, an excessive $\lambda_O$ prioritizes the hallucination around occluded parts at the expense of the amodal prediction, as shown in Fig. \ref{imgL}. Based on these results, we select $\lambda_O = 0.25$ as the optimal configuration.

\begin{table}[t]
\resizebox{\columnwidth}{!}{%
\begin{tabular}{cccc}\hline
\multirow{2}{*}{Model} & \multicolumn{3}{c}{$mIoU_O$} \\
& \multicolumn{1}{c}{\textit{$occ_R$-low} } & \multicolumn{1}{c}{\textit{$occ_R$-medium}} & \multicolumn{1}{c}{\textit{$occ_R$-high}} \\ \hline \hline
C2f-Seg & 29.57 & 76.69 & 71.46 \\
\begin{tabular}[c]{@{}c@{}}ViTA-Seg Single Head\end{tabular} & 25.82 & 76.30 & 69.39 \\
\begin{tabular}[c]{@{}c@{}}ViTA-Seg Dual Head\end{tabular} & \textbf{36.48} & \textbf{77.14} & \textbf{73.59} \\ \hline 
\end{tabular}%
} 
\caption{$mIoU_O$ on ViTA-SimData by $occ_R$.}
\label{occtab}
\end{table}

\section{Conclusion}
This work introduced ViTA-Seg, a class-agnostic ViT framework for amodal segmentation designed to handle the severe occlusions typical of robotic bin picking. By exploiting global self-attention, ViTA-Seg predicts complete object shapes, including occluded regions, without relying on category-specific priors. Two architectures were proposed: Single Head, which outputs the amodal mask, and Dual Head, which jointly predicts amodal and occluded masks through cross-task collaboration. We also developed ViTA-SimData, a photo-realistic synthetic dataset tailored to bin-picking scenarios. Using COCOA, KINS, and ViTA-SimData, we conducted extensive experiments showing that ViTA-Seg Dual Head consistently outperforms the state-of-the-art C2F-Seg. Specifically, it improves amodal and occluded prediction by 6 and 13 points on COCOA, 4 and 6 points on KINS, and 6 and 5 points on ViTA-SimData, respectively. Furthermore, ViTA-Seg achieves approximately 12 times faster inference, enabling real-time performance. Beyond bin picking, ViTA-Seg demonstrates improved performance also on the real-world datasets KINS and COCOA, indicating its suitability for broader visual perception applications involving occluded scenes. 

Future work will be aimed at evaluating ViTA-Seg directly in a real bin-picking setup to assess its performance under fully unconstrained industrial conditions. Additional future efforts will focus on collecting a dedicated real amodal dataset for bin picking, providing a valuable benchmark for advancing research on amodal perception in robotic manipulation.perception in robotic manipulation.

\begin{table}[]
\centering
\resizebox{0.7\columnwidth}{!}{%
\begin{tabular}{cccc}\hline
\multirow{2}{*}{$\lambda_O$} & \multicolumn{3}{c}{ViTA-Seg Dual Head} \\
& $mIoU_A$ & $mIoU_V$ & $mIoU_O$ \\ \hline \hline
0       & 88.31 & 91.01 & 49.33 \\ 
0.25    & \textbf{91.09} & \textbf{91.48} & 58.65 \\
0.50    & 91.00 & 91.27 & \textbf{58.98} \\ \hline
\end{tabular} 
} 
\caption{Performance metrics on ViTA-SimData for different $\lambda_O$.}
\label{lambda_O}
\end{table}

\bibliography{ifacconf}             
\end{document}